\documentclass[journal]{IEEEtran}
\usepackage{times}
\usepackage{latexsym}
\usepackage{graphicx}
\usepackage{amsmath}
\usepackage{amsthm}
\usepackage{booktabs}
\usepackage{algorithm}
\usepackage{algorithmic}
\usepackage{amssymb}
\usepackage{multirow}
\usepackage{multicol}
\usepackage{tabularx}
\usepackage{arydshln}
\usepackage{etoolbox}
\usepackage{url}
\makeatletter
\patchcmd{\@makecaption}
  {\scshape}
  {}
  {}
  {}
\makeatletter
\patchcmd{\@makecaption}
  {\\}
  {.\ }
  {}
  {}
\makeatother
\def\tablename{Table}

\ifCLASSINFOpdf
\else
\fi
\begin{document}
\title{Reference Knowledgeable Network\\ for Machine Reading Comprehension}

\author{Yilin~Zhao, Zhuosheng Zhang, Hai~Zhao
\IEEEcompsocitemizethanks{
\IEEEcompsocthanksitem{This paper was partially supported by Key Projects of National Natural Science Foundation of China (U1836222 and 61733011).}
\IEEEcompsocthanksitem Yilin~Zhao, Zhuosheng Zhang, Hai~Zhao are with the Department of Computer Science and Engineering, Shanghai Jiao Tong University, and also with Key Laboratory of Shanghai Education Commission for Intelligent Interaction and Cognitive Engineering, Shanghai Jiao Tong University, and also with MoE Key Lab of Artificial Intelligence, AI Institute, Shanghai Jiao Tong University. \protect\\ E-mail: \{zhaoyilin, zhangzs\}@sjtu.edu.cn, zhaohai@cs.sjtu.edu.cn.}}


\markboth{IEEE/ACM TRANSACTIONS ON AUDIO, SPEECH, AND LANGUAGE PROCESSING}%
{Shell \MakeLowercase{\textit{Zhao et al.}}: Reference Knowledgeable Network for Machine Reading Comprehension}



\maketitle

\begin{abstract}
Multi-choice Machine Reading Comprehension (MRC) as a challenge requires models to select the most appropriate answer from a set of candidates with a given passage and question. Most of the existing researches focus on the modeling of specific tasks or complex networks, without explicitly referring to relevant and credible external knowledge sources, which are supposed to greatly make up for the deficiency of the given passage. Thus we propose a novel reference-based knowledge enhancement model called \emph{\textbf{Re}ference \textbf{K}nowledgeable \textbf{Net}work (RekNet)}, which simulates human reading strategies to refine critical information from the passage and quote explicit knowledge in necessity. In detail, \emph{RekNet} refines fine-grained critical information and defines it as \emph{Reference Span}, then quotes explicit knowledge quadruples by the co-occurrence information of \emph{Reference Span} and candidates. The proposed \emph{RekNet} is evaluated on three multi-choice MRC benchmarks: RACE, DREAM and Cosmos QA, obtaining consistent and remarkable performance improvement with observable statistical significance level over strong baselines. Our code is available at \url{https://github.com/Yilin1111/RekNet}.
\end{abstract}


% \fi
\IEEEpeerreviewmaketitle

\section{Introduction}
\label{sec:introduction}

\IEEEPARstart{M}{achine} reading comprehension (MRC) is a challenging natural language understanding task which requires machines to answer questions according to given passages \cite{Moritz2015teaching,wang2019evidence}. 
According to the formats of expectant answers \cite{baradaran2020survey,khashabi2020unifiedqa}, MRC tasks can be roughly divided into generative task (to generate answer texts to given questions) \cite{Schwarz2018narrativeqa}, extractive task (to extract spans from given contexts for answer prediction) \cite{yang2018hotpotqa}, multi-choice task (to select the most appropriate answer among given candidates) \cite{richardson2013mctest} and Yes/No task (to identify the authenticity of given declarative sentences, which can be regarded as a simplified version of multi-choice task) \cite{clark2019boolq}.
Among above tasks, multi-choice task requires model to have stronger ability for reading and comprehension due to potential gap between contexts and given candidates, which is the focus of this work.

% Early MRC datasets usually provide passages whose contents are extracted from articles \cite{rajpurkar2018SQuAD2,lai2017race}. 
% Recently, conversational reading comprehension has aroused great interests whose passages are derived from multi-turn dialogue segments \cite{,sun2019dream}, making the task be more challenging.

\begin{figure}[htbp]
\centering
\includegraphics[scale=0.5]{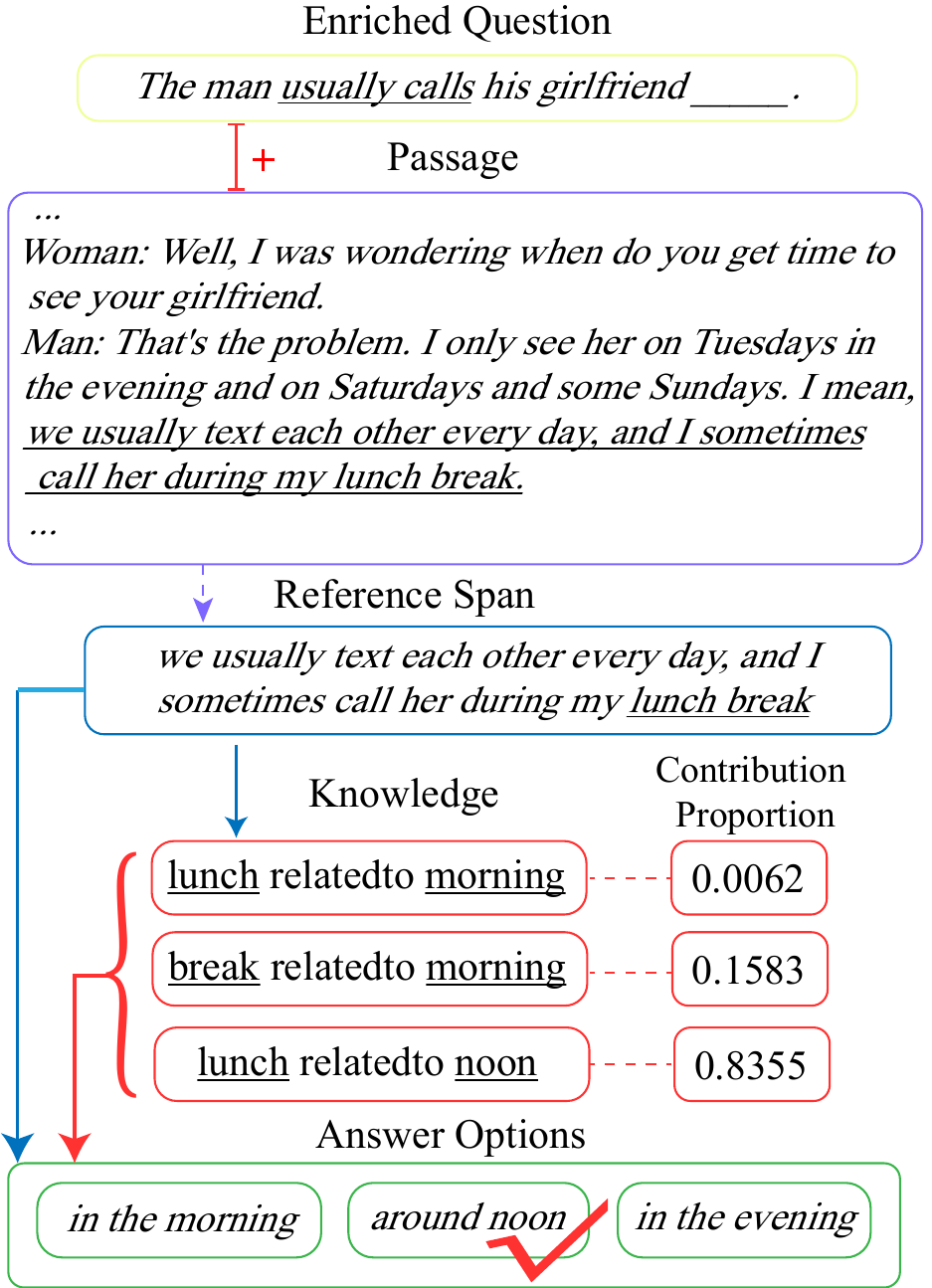}
\caption{An sample process of our model. In this example, the enriched question is the same as the original question.}
\label{example}
\end{figure}

In recent years, various datasets and tasks have been proposed, promoting a rapid improvement of MRC techniques \cite{lowe2015ubuntu,wang2018glue,reddy2019coqa}.
To solve challenging MRC tasks, one of the most popular practice to is adopting powerful pre-trained contextualized models as encoder to obtain contextualized representations \cite{peters2018elmo,devlin2019bert,lewis2020bart,radford2018gpt}.
Instead of better exploiting pre-trained encoders, this work is motivated by human reading and comprehension strategies to improve the performance of model.
According to \textit{Dual Process Theory} of cognition psychology \cite{evans1984heuristic,wason1974dual,evans2003two,kahneman2011thinking,evans2017dual}, the cognitive process of human brains potentially involves two distinct types of procedures: contextualized perception (\textit{reading}) and analytic cognition (\textit{comprehension}), where the former gathers information in an implicit process, then the latter conducts the controlled reasoning and execute goals \cite{zhang2020survey}.
In further researches, Bengio \cite{bengio2003neural} summarized two above cognitive sub-processes into more universal information processing systems: \textit{Implicit System} and \textit{Explicit System}, providing the paradigm architecture for machines to simulate human reading and comprehension strategies.

Inspired by above theories and strategies, we decouple MRC into \textit{sketchy reading} by extracting the critical spans from passages, and \textit{extensive reading} by quoting external knowledge explicitly.
Furthermore, we propose a knowledge enhancement model based on extracted critical information, called \emph{RekNet (\textbf{Re}ference \textbf{K}nowledgeable \textbf{Net}work)}.
In detail, \emph{RekNet} refines the fine-grained critical information by a span extraction module and defines it as \emph{Reference Span}, then quotes relevant explicit knowledge in the form of quadruples by the co-occurrence information of \emph{Reference Span} and candidate answers.
One sample process of \emph{RekNet} is shown in Figure \ref{example}. 
After integrating the passage and enriched question (the integration of the original question and co-occurrence information of all candidate answers), we refine \emph{Reference Span} to obtain relevant knowledge quadruples for answer prediction.

The proposed \emph{RekNet} is evaluated on three multi-choice MRC benchmarks, RACE \cite{lai2017race}, DREAM \cite{sun2019dream} and Cosmos QA \cite{huang2019CosmosQA}, with significant and consistent performance improvement compared to baselines, and passes the significance test of MRC task \cite{zhang2020retro}.
In summary, our main contributions are follows:\\
i) We propose a novel reference-based knowledge enhancement model \emph{RekNet}, which makes the first attempt to integrate critical information extraction and knowledge injection for inference on MRC tasks.\\
ii) \emph{RekNet} uses novel knowledge quadruples to quote relevant and credible knowledge, highlighting the effectiveness of quantitative knowledge items in multi-choice MRC tasks.

\section{Related Studies}
To solve multi-choice MRC tasks, existing studies mostly focus on more powerful pre-trained contextualized encoders \cite{devlin2019bert,lan2019albert,yang2019xlnet,clark2019electra}, or more complex network design to model the interactions between input sequences \cite{zhang2020semBERT,zhu2020dual,wang2018co,tang2019multi}.
Instead, some studies attempt to improve the performance by human reading strategy simulation, among which \emph{knowledge injection} and \emph{reading strategy simulation} are in widespread attention.

\subsection{Knowledge Injection}
Observing the drawbacks of the lack of commonsense in MRC models, some researchers attempt to inject external knowledge explicitly \cite{lin2021differentiable,Bhagavatula2020Abductive,shwartz2020unsupervised}.
Xia \cite{xia2019auxiliary} used auxiliary tasks to obtain relevant knowledge, while Mihaylov \cite{mihaylov2018knowreader} encoded external commonsense knowledge as key-value memory under cloze-style setting.
For non-MRC tasks, Chaudhuri \cite{Chaudhuri2018domain-knowledge} quoted domain-specific knowledge with GRUs to refine and encode keywords, and Lin \cite{lin2019kagnet} quoted a conceptual sub-graph of external knowledge to get better performance.
Feng \cite{feng2020mhgrn} quoted multi-hop commonsense for simple questions without relevant contexts, while Lin \cite{lin2020commongen} proposed a constrained text generation task for generative commonsense reasoning.
With the increasing negative impact of knowledge noise on model performance, researchers also explored to quote more relevant knowledge and reduce knowledge noise \cite{kim2020skt,liu2019kbert}.
Though some of the above studies have noticed the importance of relevant knowledge selection, they generally ignore the powerful utility of critical information to knowledge injection as well as the filtration of untrustworthy knowledge.

\subsection{Reading Strategy Simulation}
Inspired by human reading strategies, some researchers design specific reading strategies for the model to improve MRC performance \cite{li2018unified}. 
Zhang \cite{zhang2020retro} proposed a retrospective reader for span-based MRC tasks, and Sun \cite{sun2019strategy} designed three specific human reading strategies for modeling enhancement.
Among all reading strategies, \emph{evidence information extraction} has been paid great attention \cite{choi2017coarse}.
Wang \cite{wang2019evidence} gave the first attempt to extract evidence sentences in single multi-choice MRC task, and Yadav \cite{yadav2019quick} applied this method to multi-hop QA tasks.
Niu \cite{niu2020soft-evidence} supervised evidence extractor with auto-generated evidence labels in an iterative process. 
However, all the above studies only execute information extraction at a more coarse-grained (sentence-level) granularity, lacking of sufficient accuracy compared to information extraction at fine-grained granularity, as well as adequate interaction between salient knowledge pieces.

\subsection{Our Method}
This work differs from previous studies as following:\\
i) To highlight question-aware critical information from context, we model the \emph{Reference Span}, instead of using the whole context for inference or explicit knowledge retrieving;\\
ii) To alleviate the negative impact of untrustworthy knowledge, our method quotes relevant explicit knowledge in the form of quadruple by adding confidence value of each knowledge items as an quantitative indicator over previous knowledge triplet.\\
To our best knowledge, our model is the first \emph{reference-based knowledge enhancement} model for multi-choice MRC tasks.

\section{Preliminary Experiments}

To explore general characteristics of multi-choice MRC tasks, we randomly extract $50$ examples in DREAM, RACE and Cosmos QA respectively, finding that $38\%, 28\%$ and $34\%$ of the examples can be inferred \textbf{just} by several \emph{adjacent phrases} in one single sentence, without explicit knowledge or logical calculation.
This finding indicates that, in multi-choice MRC tasks, critical fine-grained information such as concise spans may contain the salient information for answer prediction directly, which can serve as an important reference indicator.
% and the need of multi-hop inference in extensive MRC benchmarks is limited.

\begin{table}[htbp]
\caption{\label{preliminary_result} Preliminary Experiment 1: the exploration of overall utility of \textit{Reference Span}.}
	\centering{
		\begin{tabular}{p{4cm}|l|l}
		    \hline	\bf Model & \bf Dev & \bf Test  \\
			\hline \hline
			Baseline (ALBERT$_{base}$) & 65.74 & 65.56 \\
			+ Reference Span & 67.65 & 67.86\\
			\hline
			Reference Span only & 59.02 & 58.94 \\
			\hline
		\end{tabular}
	}
\end{table}

Inspired by above finding, we directly replace the original passage with extracted \emph{Reference Span} as the input on DREAM, and the replaced model achieves acceptable performance as Table \ref{preliminary_result} shows.
In detail, the \emph{Reference Span} is extracted by another ALBERT$_{base}$ pre-trained on SQuAD 2.0, and the \emph{Baseline} is fed with the triple of \{\textit{passage, question, candidate answers}\} while \emph{Reference Span only} is fed with \{\textit{Reference Span, question, candidate answers}\}.
\emph{+ Reference Span} represents the pooled logits of two above sequence triplets are concatenated for prediction, to both take advantage of the original redundant and refined critical information.
From the results of Preliminary Experiment 1, there are two main factors which may cause the unsatisfactory performance according to our observation:

i) difficult questions that require external knowledge (as Figure \ref{example} shows);

ii) potential mistakes from critical information extraction. 

We find that $26\%$ examples of DREAM, $20\%$ examples of RACE and $62\%$ examples of Cosmos QA requires explicit knowledge for answer prediction in above $50$ random examples, the first issue can be handled by augmentation from extra sources.
Furthermore, the second issue can be alleviated by modeling both the original passage-aware and refined reference-aware sequences, as the superior result ($+2.30\%$) of \emph{+ Reference Span} showing in Table \ref{preliminary_result}.

To further verify whether fine-grained information (i.e., several adjacent phrases in the preliminary finding) has more positive influence than coarse-grained information (i.e., evidence sentences in existing studies) for critical information extraction, we design Preliminary Experiment 2, where two more baselines obtain more coarse-grained \textit{Reference Sentences} on DREAM.
The first baseline calculates TF-IDF scores of each sentence in the passage and given question, denoting the largest appearing score as $s_{max}$.
Then the model retains all sentences whose scores $s$ satisfying: $s \ge 0.7 \times s_{max}$ and splices them by spaces, as the input \emph{Reference Sentence}.
The other one uses our pre-trained ALBERT$_{base}$ (with the same setting to Preliminary Experiment 1) to obtain the whole sentence that contains \textit{Reference Span} as \emph{Reference Sentence}.

\begin{table}[htbp]
	\caption{\label{fine-grained_result} Preliminary Experiment 2: the exploration of fine-grained information.}
	\centering{
		\begin{tabular}{p{4cm}|l|l}
		    \hline	\bf Method & \bf Dev & \bf Test  \\
			\hline \hline
			TF-IDF Method & 56.60 & 55.44 \\
			ALBERT$_{base}$ Method (sentence) & 58.67 & 58.35 \\
			ALBERT$_{base}$ Method & 59.02 & 58.94 \\
			\hline
		\end{tabular}
	}
\end{table}

As Table \ref{fine-grained_result} shows, with \emph{Reference Span} we extracted, the baseline obtains extra $0.47\%$ improvement on average comparing with general coarse-grained information extraction method, which proves the superiority of the fine-grained \emph{Reference Span}.

Inspired by both our findings above and human reading and comprehension experiences \cite{ding2019cognitive,zhang2020retro}, we design our model by following a two-stage reading strategy, as \emph{critical information extraction} and \emph{knowledge injection}:
% In the human reading pattern, one will highlight critical information for the given question in a sketchy reading first, then integrate extensive relevant knowledge sources to extrapolate the appropriate answer. 

i) In the task perspective, the machine should focus on question-relevant information from the lengthy passage, which can interpret the process of human reading comprehension.
We call it \emph{sketchy reading}, which embodies \emph{reading} process in MRC.

ii) In the model perspective, the machine should solve the given questions with transcendental external knowledge and the current context.
We call it \emph{extensive reading}, which embodies \emph{comprehension} process in MRC.

% In \emph{extensive reading}, due to MRC tasks are rich in text content and model may quote a large amount of irrelevant or untrustworthy knowledge, we quoted knowledge in the form of quadruples to filter distracting information\footnote{The comparison to triplets used in previous studies is shown in Section 6.2.}.

\section{RekNet}
\begin{figure}[htbp]
\centering
\includegraphics[scale=0.8]{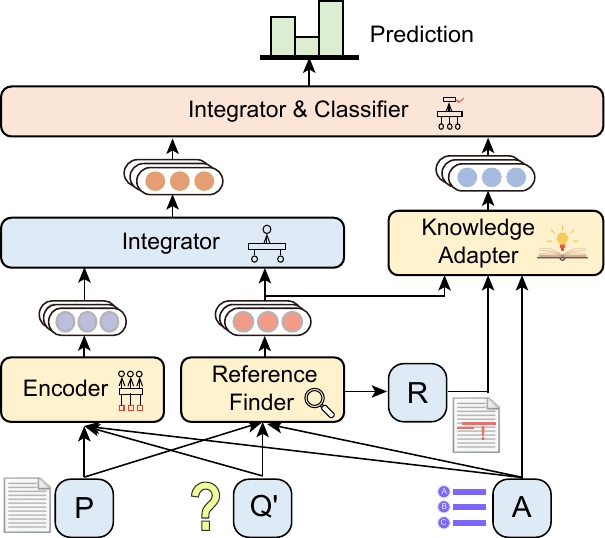}
\caption{Overview of \emph{RekNet}. $P$, $Q'$, $A$, $R$ stand for \emph{Passage}, \emph{Enriched Question}, \emph{Candidate Answers} and \emph{Reference Span} respectively.}
\label{model}
\end{figure}
Multi-choice MRC tasks can be defined as a triplet $(P, Q, A)$, where $P$ is passage, $Q$ is question with $n$ candidate answers: $A$ = $\{A_1, ..., A_n\}$.
Let $A_{correct} \in$ $A$ be the correct answer of $Q$, the aim of multi-choice MRC model is to make:
$$\mathcal{P}(A_{correct} \mid P, Q, A) \ge \mathcal{P}(A_i \mid P, Q, A),$$
where $i \in (1, ..., n)$ and $\mathcal{P}$ represents the probability of each candidate answer.

Another element $R$ is employed in our proposed \emph{RekNet}, which is implemented as the critical information span (named as \emph{Reference Span}) for each question.
Furthermore, we combine the co-occurrence information of all candidate answers $A$ with the original question $Q$, generating enriched question $Q'$ for \emph{RekNet}. 
Therefore, \emph{RekNet} defines multi-choice MRC tasks to a quadruple $(P, Q', R, A)$ instead.

The overall framework of \emph{RekNet} is shown in Figure \ref{model}, consisting of three modules: \emph{Reference Finder}, \emph{Knowledge Adapter} and \emph{Integrator}.
Triplet $(P, Q', A)$ is the input of \emph{RekNet}, and the lower \emph{Integrator} simulates human \emph{sketchy reading} while the upper \emph{Integrator} simulates human \emph{extensive reading}.
Details of above modules are shown in following subsections.

\subsection{Reference Finder}
\begin{figure}[htbp]
\centering
\includegraphics[scale=0.9]{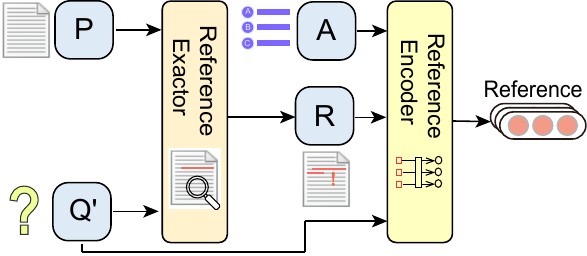}
\caption{The framework of \textit{Reference Finder}.}
\label{reference}
\vspace{-0.3cm}
\end{figure}

\emph{RekNet} employs \textit{Reference Finder} to extracts \emph{Reference Span}\footnote{In this work, \emph{RekNet} only extracts one consequent \emph{Reference Span}, which leads to the optimal performance in evaluated multi-choice MRC benchmarks.} and encodes it for critical information enhancement and relevant knowledge filtration.
\textit{Reference Finder} has two main components, called \textit{Reference Extractor} and \textit{Reference Encoder}, as Figure \ref{reference} shows.

\noindent$\bullet$ \textbf{Enriched Question Q'}\\
One challenge in multi-choice MRC task is the implicit critical information in given candidate answers. 
Without referring to candidate answers, it may be hard for model to extract precise critical information for the given question.
Take Dialogue 1 in Table \ref{enriched} for example, only model knows the key point to the given question is \emph{working situation of the man}, it can refine the critical information accurately.
Thus, for each question, we pick up the co-occurrence information of all candidate answers and add them to the head of the original question $Q$ in order, forming the enriched question $Q'$.
Above action facilitates \textit{Reference Finder} to extract \emph{Reference Span} more precisely.

\begin{table}[htbp]
\caption{\label{enriched} Sample dialogue in DREAM for $Q'$, where \textbf{bold texts} in passage represent critical information for answer prediction while those in answer candidates represent co-occurrence information for the generation of $Q'$.}
\centering
\begin{tabular}{|p{8.2cm}|}
\hline \bf Dialogue 1\\\hline
...\\
\emph{W: \textbf{You worked for a large company before}, didn't you?}\\
\emph{M: Yes, I did. \textbf{But I prefer a small company}.}\\
\emph{W: Is it really different?}\\
\emph{M: Oh, yes. It's much different. I like a small company because it's more exciting.}\\
...\\\hline
Q: \emph{What do we learn from the conversation?}\\
A. \emph{\textbf{The man} has been \textbf{working in a small company} for a long time.}\\
B. \emph{\textbf{The man} used to work for a big \textbf{company}, but now he \textbf{works in a small} one.} (correct)\\
C. \emph{\textbf{The man works in a small company}, but he doesn't like it.}\\\hdashline
Q': \emph{The man work in a small company. What do we learn from the conversation?}\\\hline
\end{tabular}
\end{table}

In detail, for each candidate answer $A_i \in A$, we construct a prototype token set $\mathcal{A}_i$ which contains the prototypes of all tokens in $A_i$.
We employ $spaCy$\footnote{\url{https://spacy.io/}} to execute lemmatization of each token in $A_i$, and obtain all prototypes to constitute above prototype token set $\mathcal{A}_i$.
Then we keep a co-occurrence token set $T=\{t_1,...,t_c\}$, where $c$ is the total of co-occurrence tokens, and for each co-occurrence token $t_j \in T$ we have:
$$t_j \in \mathcal{A}_i, \quad \forall i \in \{1,...,n\}.$$
We splice all prototype tokens in $T$ with space in order and add the co-occurrence information with ``.” into the head of $Q$ to constitute $Q'$ ultimately.

\noindent$\bullet$ \textbf{Reference Extractor}\\
\textit{RekNet} employs a pre-trained contextualized encoder as \textit{Reference Extractor} to extract \emph{Reference Span}.
Following \cite{devlin2019bert}, the input sequence of \textit{Reference Extractor} is $[CLS]\;Q'\;[SEP]\;P\;[SEP]$.
We define the hidden size of the pre-trained contextualized encoder as $H$ and the representation on the final hidden layer for the $i$-th input token as $T_i \in \mathbb{R}^H$, then we score the predicted span starts from the $i$-th input token and ends at the $j$-th input token as $S \cdot T_i+E \cdot T_j$, where $S/E\in \mathbb{R}^H$ is the introduced start/end vector. 
Ultimately, span with the largest score of the former formula is chosen as the predicted \emph{Reference Span}.

\noindent$\bullet$ \textbf{Reference Encoder}\\
\textit{RekNet} employs the contextualized encoder in its baseline model as \textit{Reference Encoder}, which shares all parameters to the \textit{Encoder} in Figure \ref{model}.
\textit{Reference Encoder} encodes each $(Q',A_i)$ pair with $R$ in the form of $[CLS]\;R\;[SEP]\;Q'+A_i\;[SEP]$\footnote{$+$ denotes string splicing operation by one space.}, making \emph{RekNet} obtain critical information enhancement from \emph{Reference Span} directly;
and encodes out a set of \textit{Reference Vectors} $RV = \{RV_1,..., RV_n\}$ where $RV_i \in \mathbb{R}^H, i \in \{1, ..., n\}$\footnote{$RV_i$ is the embedding vector of $[CLS]$ from the last hidden layer of \textit{Reference Encoder}.}.

Similar to \textit{Reference Encoder}, \textit{Encoder} in Figure \ref{model} takes $[CLS]\;P\;[SEP]\;Q'+A_i\;[SEP]$ as its input sequence, and produces a similar set of \textit{Passage Vectors} $PV = \{PV_1,..., PV_n\}$, where $PV_i \in \mathbb{R}^H, i \in \{1, ..., n\}$. 
Ultimately, each element $RV_i$ in $RV$ will fuse with corresponding $PV_i$ in the lower \emph{Integrator} in Figure \ref{model}.

\subsection{Knowledge Adapter}
\begin{figure}[htbp]
\centering
\includegraphics[scale=0.8]{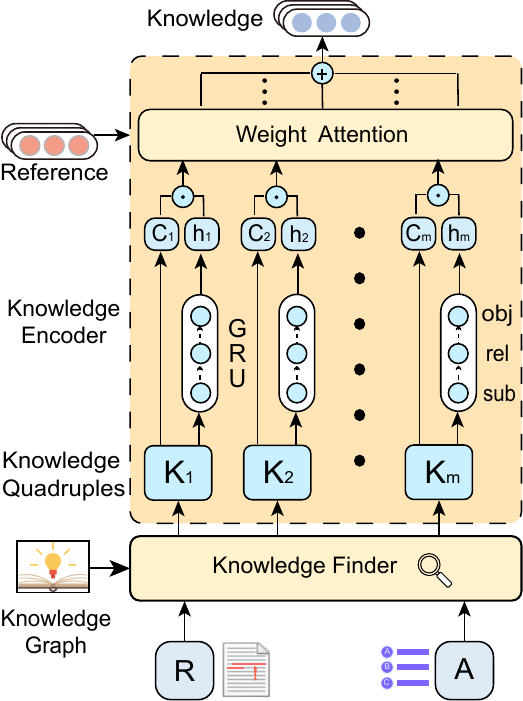}
\caption{The framework of \textit{Knowledge Adapter}. \textbf{·} stands for scalar multiplication operation, and $m=k \times n$ is the number of knowledge quadruples. $K_j$ refers to the $j$-th knowledge quadruple, $C_j$ and $h_j$ refers to the confidence value and embedding representation of $K_j$.}
\label{knowledge}
\vspace{-0.3cm}
\end{figure}

\emph{RekNet} employs \textit{Knowledge Adapter} to quote relevant explicit knowledge and encode knowledge quadruples, to highlight credible knowledge items.
Figure \ref{knowledge} depicts the framework of \textit{Knowledge Adapter}, which is composed of \textit{Knowledge Finder} and \textit{Knowledge Encoder}.

\noindent$\bullet$ \textbf{Knowledge Finder}\\
For each input example, \emph{Knowledge Finder} searches question-relevant knowledge items based on $(R,A)$ binary, due to the plentiful information and relatively concise length of $R$.
% \textit{Knowledge Finder} searches knowledge quadruples for each $(P,Q,A)$ triplet.
% Due to passage $P$ usually has a lengthy context which contains abundant noise, \emph{RekNet} sends $(R,A)$ binaries as input instead.
In detail, \textit{Knowledge Finder} quotes knowledge items whose \emph{subject entity} and \emph{object entity} appear in the prototype list of $R$ and $A$ respectively.
Above prototype matching is similar to the prototype matching for enriched question generation: we employ $spaCy$ to execute lemmatization of each token in $R$ and $A$, and obtain prototypes to constitute prototype lists of $R$ and $A$.
Besides, we also attempt cosine similarity for entity matching, getting no significant improvement compared with above simple method.

Knowledge items are saved in the form of quadruples $(sub, rel, obj, con)$, which stands for the subject, relation, object and confidence value of each knowledge item.
\emph{(doctor, can, help\_sick\_person, 4.472)} is a sample quadruple, where the subject, relation, and object can be a word or a phrase, and confidence value is a number larger than $0.1$.
Under this setting, \emph{RekNet} can effectively separate knowledge quadruples with low confidence values from blank knowledge quadruples.
Larger confidence value indicates the knowledge item is more credible, and we obtain confidence values from the \textit{weight values} in ConceptNet \cite{speer2017conceptnet} then normalize them into $[0.1, 2.0]$, by setting the confidence values larger than $2.0$ to $2.0$ due to their minor scale and uneven distribution.
One typical \textit{weight value} in the original ConceptNet is $1.0$, but the value can be higher or lower, and $99.37\%$ of the \textit{weight values} locate in $[0.1, 2.0]$ in the original ConceptNet.

\textit{RekNet} sets each input example can only retain $k \times n$ knowledge quadruples with the largest confidence values at most, where $k$ is the number of quotable knowledge quadruples for each candidate answer, and this design can reduce knowledge noise significantly.
If one example can not get enough quadruples, \textit{RekNet} will add blank quadruples whose confidence values are $0$ to fill up to the specified amount.

\noindent$\bullet$ \textbf{Knowledge Encoder}\\
We define the remaining part of \textit{Knowledge Adapter} as \textit{Knowledge Encoder}. 
For each $(R,A_i)$ binary, the input embedding for each word or phrase in knowledge quadruples is sent to a series of GRU modules to encode the quadruple as following:
$$hidden_{sub} = GRU(emb_{sub}, 0), $$
$$hidden_{rel} = GRU(emb_{rel}, hidden_{sub}), $$
$$hidden_{obj} = GRU(emb_{obj}, hidden_{rel}), $$
where $emb_{sub}$, $emb_{rel}$, $emb_{obj}$ are the input embeddings of subject, relation and object, and $hidden_{sub}$, $hidden_{rel}$, $hidden_{obj}$ are the last hidden representations of GRU.
With above encoding design, the directionality of input quadruples can be retained, and the embedding of explicit knowledge is in the same vector space as the plain tokens. 

Then \textit{RekNet} sends all $m=k\times n$ embeddings to a weighted attention module with their confidence values, and the main operation can be expressed as:
$$KV_i=\sum_{j=1}^{k\times n}Softmax(WeiAtt(RV_i, h_j, c_j))^Th_j,\;i\in\{1,...,n\},$$
where $h_j$ and $c_j$ are $hidden_{obj}$ and confidence value of the $j$-th knowledge quadruple, $KV_i$ and $RV_i$ are respectively \textit{Knowledge Vector} and \textit{Reference Vector} of the $(R,A_i)$ binary.
The weighted attention module $WeiAtt$ is employed to compute the unnormalized \textit{contribution proportion} for each knowledge quadruple, which can be formulated as:
$$WeiAtt(RV_i, h_j, c_j)=(RV_i^TWh_j) * c_j,\;j\in \{1,...,m\},$$
where $W$ is a linear transform matrix, and $*$ is scalar multiplication.

\subsection{Integrator \& Classifier}
\begin{figure}[htbp]
\centering
\includegraphics[scale=0.8]{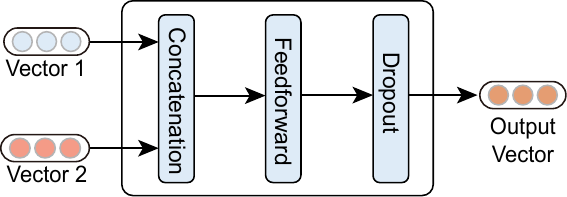}
\caption{The framework of \textit{Integrator}, where all vectors are in the size of $\mathbb{R}^H$.}
\label{integration}
\end{figure}

\noindent$\bullet$ \textbf{Integrator}\\
There are two \textit{Integrators} in \textit{RekNet} to integrate information from two input embedding representations to one fusion representation.
The structure of \textit{Integrator} is illustrated in Figure \ref{integration}.

In detail, there are three layers in \textit{Integrator}. 
In Concatenation Layer, two input embedding vectors in size of $\mathbb{R}^H$ are spliced into a vector in size of $\mathbb{R}^{2\times H}$. 
Then the dimension of spliced vector is reduced to $\mathbb{R}^H$ in Feedforward Layer, which is a linear layer.
Ultimately, there is a Dropout Layer to prevent over-fitting.

\noindent$\bullet$ \textbf{Classifier}\\
For classifier, \textit{RekNet} employs a linear softmax layer to calculate score for each candidate answer.
Furthermore, we use standard Cross Entropy Loss as the loss function, which is the same as the mainstream works in MRC field.

\section{Experiments}
\subsection{Setup}
We run the experiments on $8$ NVIDIA Tesla P40 GPUs, and the implementation of \textit{RekNet} is based on the Pytorch implementation of ALBERT$_{xxlarge}$ \cite{lan2019albert}. 
In the experiments, we adopt ALBERT pre-trained on SQuAD 2.0 \cite{rajpurkar2018SQuAD2} as \emph{Reference Extractor}, due to SQuAD 2.0 has the largest contribution to other MRC tasks among extractive tasks \cite{khashabi2020unifiedqa}.
Besides, we eliminate the possibility of extracting null \emph{Reference Span}\footnote{We eliminate above possibility by drastically increasing the threshold $\tau$ in \textit{Reference Extractor}. According to \cite{devlin2019bert}, when $S \cdot T_0+E \cdot T_0 > \max_{i\leq j} S \cdot T_i+E \cdot T_j + \tau$, \emph{Reference Extractor} will return a null \emph{Reference Span}.}, to ensure all examples can get a non-null \emph{Reference Span}, because there exist questions with null span as the golden answer in SQuAD 2.0.
We set the number of knowledge quadruples for each candidate answer \emph{k} to $5$, which leads to the best performance for \textit{RekNet}.
The fine-tuning hyper-parameters of \emph{RekNet} are given in Table \ref{hyperparameter}, which leads to the optimal performance.
In detail, statistics about the length are based on token level, and \textit{Source Content} denotes the input sequence of model ($[CLS]\;R\;[SEP]\;Q'+A_i\;[SEP]$).

\begin{table}[htbp]
\caption{\label{hyperparameter} The fine-tuning hyper-parameters of \emph{RekNet}.}
	\centering{
		\begin{tabular}{p{4cm}|l|l|p{1cm}}
		    \hline	\bf Hyper-parameter & \bf RACE & \bf DREAM & \bf Cosmos QA  \\
			\hline \hline
			Learning Rate & 1e-5 & 1e-5 & 1e-5 \\
			Batch Size & 32 & 24 & 32\\
			Warmup Steps & 1000 & 50 & 1000\\
			Maximum Sequence Length & 384 & 512 & 384\\
			Maximum \emph{Reference Span} Length & 256 & 256 & 256\\
			Training Epochs & 2 & 2 & 5\\
			Steps to Save Checkpoints & 3000 & 400 & 2000\\\hdashline
			Maximum Source Content Length & 1480 & 1425 & 396\\
			Average Source Content Length & 365 & 176 & 113\\
			\hline
		\end{tabular}
	}
\end{table}

\subsection{Dataset}
We employ RACE \cite{lai2017race}, DREAM \cite{sun2019dream} and Cosmos QA \cite{huang2019CosmosQA} as our evaluation benchmarks, and ConceptNet 5.7.0 \cite{speer2017conceptnet} as external commonsense knowledge source.
The details of above datasets are shown as following:

\textbf{RACE} is a large-scale MRC task collected from English examinations, which contains nearly 100,000 questions.
The passages are in the form of articles with diversified topic domains, and most questions require contextual reasoning.
Each question in RACE has 4 candidate answers.

\textbf{DREAM} is a dialogue-based dataset for multi-choice MRC, containing more than 10,000 questions.
More than 80\% of the questions in DREAM are non-extractive and require reasoning from multi-turn dialogues, and more than one third of the given questions involve commonsense knowledge.
Each question in DREAM has 3 candidate answers.

\textbf{Cosmos QA} is a large-scale MRC task collected from people’s daily narratives, which has about 35,600 questions.
The questions are about the causes or effects of events, which require evidence extraction as well as knowledge injection.
Each question in Cosmos QA has 4 candidate answers.

\textbf{Knowledge Source}
ConceptNet 5.7.0 \cite{speer2017conceptnet} is one of the largest structured knowledge base with confidence weight for each knowledge item as its knowledge source. 
Due to the huge size of ConceptNet, we only retain its English concepts.
Following Lin \cite{lin2019kagnet}, we merge some relations which share similar meanings, and the retained relations are listed in Table \ref{relations}.
To obtain the input embedding for each word or phrase in ConceptNet, we adpote the pre-trained $100$-dimension GloVe \cite{pennington2014glove} embedding vectors.

\begin{table}[htbp]
\caption{\label{relations} Retained relations in ConceptNet.}
\centering
\begin{tabular}{|p{8.2cm}|}
\hline \bf Retained relations in ConceptNet\\\hline
locate, can, causes, product, desires, antonym, situation, is, entails, isa, disable, unnecessary, have, relatedto, field, capital, influence, occupation, language, leader\\\hline
\end{tabular}
\end{table}

\subsection{Results}
% \begin{table*} [htbp]
% 	\centering{
% 		\begin{tabular}{l|c c|c c|c c}
% 			\hline\multirow{2}{*}{\textbf{Model}} & \multicolumn{2}{c|}{\bf DREAM} & \multicolumn{2}{c|}{\bf RACE} & \multicolumn{2}{c}{\bf Cosmos QA}\\
% 			~ & \bf Dev & \bf Test & \bf Dev(M/H) &\bf Test(M/H) & \bf Dev & \bf Test\\\hline\hline
% 			BERT$_{base}$ & 63.4 & 63.2 & 64.6 (-- / --) & 65.0 (71.1 / 62.3) & 66.2 & 67.1\\
% 			BERT$_{large}$ & 66.0 & 66.8 & 72.7 (76.7 / 71.0) & 72.0 (76.6 / 70.1) & -- & --\\
% 			XLNet$_{large}$ & -- & 72.0 & 80.1 (-- / --) & 81.8 (85.5 / 80.2) & -- & --\\
% 			RoBERTa$_{large}$ & 85.4 & 85.0 & -- (-- / --) & 83.2 (86.5 / 81.8) & 81.7 & 82.7\\
% 			ALBERT$_{xxlarge}$ & 89.2 & 88.5 & -- (-- / --) & 86.5 (89.0 / 85.5) & -- & 84.9\\\hline
% 			ALBERT$_{base}$ (rerun) & 65.7 & 65.6 & 67.9 (72.3 / 65.7) & 67.2 (72.1 / 65.2) & 63.1 & 63.7\\
% 			RekNet on ALBERT$_{base}$ & 68.9 & 68.7 & 70.7 (73.9 / 68.9) & 70.3 (74.4 / 68.3) & 65.7 & 65.8\\\hline
% 			ALBERT$_{xxlarge}$ (rerun) & 88.7 & 88.3 & 86.6 (89.4 / 85.2) & 86.5 (89.2 / 85.4) & 85.0 & 84.8\\
% 			RekNet on ALBERT$_{xxlarge}$ &\bf 89.8 &\bf 87.8 (91.1 / 86.4) &\bf 87.8 (90.1 / 86.8) &\bf 85.9 &\bf 85.7\\\hline
% 		\end{tabular}
% 	}
% 	\caption{\label{result} Results on DREAM, RACE and Cosmos QA. Results in the first domain are from the leaderboards.}
% \end{table*}

We adopt accuracy as evaluation criteria for multi-choice MRC.
Tables \ref{dream_result}-\ref{cosmos_result} show the results of \textit{RekNet} compared with its baselines, as well as other public models on the leaderboard.
Though some public works may have better performance with larger model size (e.g., $3.9$ billion parameters for Megatron-BERT, $11$ billion parameters for T5) or special complex neural network (like DUMA), our proposed \textit{RekNet} obtains consistent improvements over all evaluated benchmarks, with acceptable model size.
For quantitative proof, significant test \cite{zhang2020retro} shows that \textit{RekNet} improves its baseline on evaluated benchmarks significantly with $p-value < 0.01$.

\begin{table} [htbp]
\caption{\label{dream_result} Results on DREAM. Results in the first domain are from the leaderboard. MTL denotes multi-task learning.}
	\centering{
		\begin{tabular}{p{5cm}|c|c}
			\hline\bf Model & \bf Dev & \bf Test \\\hline\hline
			FTLM++ \cite{sun2019dream} & 58.1 & 58.2    \\
			BERT$_{base}$ \cite{devlin2019bert} & 63.4 & 63.2  \\
			BERT$_{large}$ \cite{devlin2019bert} & 66.0 & 66.8   \\
			XLNet$_{large}$ \cite{yang2019xlnet} & --  & 72.0 \\
			RoBERTa$_{large}$ \cite{liu2019roberta} & 85.4 & 85.0  \\
			RoBERTa$_{large}$ + MMM \cite{jin2019mmm} & 88.0 & 88.9 \\
			ALBERT$_{xxlarge}$ + DUMA \cite{zhu2020dual} & 89.9 & 90.4 \\
			ALBERT$_{xxlarge}$ + DUMA + MTL & -- & 91.8 \\\hline
			ALBERT$_{base}$ (rerun) & 65.7 & 65.6 \\
			RekNet on ALBERT$_{base}$ & 68.9 & 68.7 \\\hline
			ALBERT$_{xxlarge}$ (rerun) & 89.2 & 88.5 \\
			RekNet on ALBERT$_{xxlarge}$ & 89.8 & 89.6 \\\hline\hline
		\end{tabular}
	}
\end{table}

\begin{table} [htbp]
\caption{\label{race_result} Results on RACE. Results in the first domain are from the leaderboard. SC denotes single choice and TL denotes transfer learning.}
	\centering
	{
		\begin{tabular}{p{3.7cm}|c|c}
			\hline\bf Model & \bf Dev (M / H) & \bf Test (M / H)\\\hline\hline
			BERT$_{base}$ \cite{devlin2019bert} & 64.6 (-- / --) & 65.0 (71.1 / 62.3)  \\
			BERT$_{large}$ \cite{devlin2019bert} & 72.7 (76.7 / 71.0) & 72.0 (76.6 / 70.1)  \\
			XLNet$_{large}$ \cite{yang2019xlnet} & 80.1 (-- / --) & 81.8 (85.5 / 80.2)\\
			XLNet$_{large}$ + DCMN+ \cite{zhang2020dcmn} & -- (-- / --) & 82.8 (86.5 / 81.3)\\
			RoBERTa$_{large}$ \cite{liu2019roberta} & -- (-- / --) & 83.2 (86.5 / 81.8)\\
			RoBERTa$_{large}$ + MMM \cite{jin2019mmm} & -- (-- / --) & 85.0 (89.1 / 83.3)\\
			T5-11B \cite{raffel2020t5} & -- (-- / --) & 87.1 (-- / --)\\
			ALBERT$_{xxlarge}$ + DUMA \cite{zhu2020dual} & 88.1 (-- / --) & 88.0 (90.9 / 86.7)\\
			T5-11B + UnifiedQA \cite{khashabi2020unifiedqa} & -- (-- / --) & 89.4 (-- / --)\\
			Megatron-BERT-3.9B \cite{shoeybi2019megatron} & -- (-- / --) & 89.5 (91.8 / 88.6)\\
			ALBERT$_{xxlarge}$ +SC +TL \cite{jiang2020single} & -- (-- / --) & 90.7 (92.8 / 89.8) \\\hline
            ALBERT$_{base}$ (rerun) & 67.9 (72.3 / 65.7) & 67.2 (72.1 / 65.2) \\
			RekNet on ALBERT$_{base}$  & 70.7 (73.9 / 68.9) & 70.3 (74.4 / 68.3) \\\hline
            ALBERT$_{xxlarge}$ (rerun) & 86.6 (89.4 / 85.2) & 86.5 (89.2 / 85.4) \\
			RekNet on ALBERT$_{xxlarge}$ & 87.8 (91.1 / 86.4) & 87.8 (90.1 / 86.8)\\\hline\hline
		\end{tabular}
	}
\end{table}

\begin{table}[htbp]
\caption{\label{cosmos_result} Results on Cosmos QA. Results in the first domain are from the leaderboard, and only public works are listed.}
	\centering{
		\begin{tabular}{p{5cm}|c|c}
		\hline
			\bf Model & \bf Dev & \bf Test\\\hline\hline
			BERT$_{base}$ \cite{liu2019roberta} & 66.2 & 67.1\\
			RoBERTa$_{large}$ \cite{liu2019roberta} & 81.7 & 83.5\\
			RoBERTa$_{large}$ + CEGI \cite{liu2020cegi} & 83.8 & 83.6\\
			ALBERT$_{xxlarge}$ + GDIN \cite{tian2020gdin} & -- & 84.5\\
			RoBERTa$_{large}$ + ALICE \cite{pereira2020alice} & 83.6 & 84.6\\
			T5-11B \cite{raffel2020t5} & -- & 90.3\\
			T5-11B + UNICORN \cite{Lourie2021unicorn} & -- & 91.8\\\hline
			ALBERT$_{base}$ (rerun) & 63.1 & 63.7 \\
			RekNet on ALBERT$_{base}$ & 65.7 & 65.8 \\\hline
            ALBERT$_{xxlarge}$ (rerun) & 85.0 & 84.8 \\
			RekNet on ALBERT$_{xxlarge}$ & 85.9 & 85.7\\\hline\hline
		\end{tabular}
	}
\end{table}

Through further observation, we find the improvement of \emph{RekNet} becomes less as the size of baseline increases (e.g., $3.1\%$ v.s. $1.1\%$ on DREAM). 
By analyzing $50$ randomly selected error cases over the baselines on DREAM, the proportion of the cases can be inferred just by several adjacent phrases decreases from $38\%$ to $26\%$ on ALBERT$_{base}$ and $16\%$ on ALBERT$_{xxlarge}$, and the cases requiring external knowledge decreases from $26\%$ to $22\%$ on ALBERT$_{base}$ and $16\%$ on ALBERT$_{xxlarge}$. 
This statistics indicates that, with more parameters and stronger encoding ability, larger model might have stronger ability to capture the relationships among adjacent phrases as well as encode knowledge implicitly by itself, which weakens the original benefit of \emph{critical information extraction} and \emph{knowledge injection}.

\begin{table}[htbp]
\caption{\label{parameter} Training parameters in \textit{RekNet} and partial public works in Tables \ref{dream_result}-\ref{cosmos_result}. The rest works do not provide the number of parameters in their models.}
	\centering{
		\begin{tabular}{p{4cm}|c}
		    \hline	\bf Model & \bf Parameters \\
			\hline \hline
			ALBERT$_{base}$ \cite{lan2019albert} & 12M\\
			ALBERT$_{base}$ + DUMA \cite{zhu2020dual} & 13.5M\\
			ALBERT$_{base}$ + DCMN+ \cite{zhang2020dcmn} & 19.4M\\
			BERT$_{base}$ \cite{devlin2019bert} & 108M\\
            ALBERT$_{xxlarge}$ \cite{lan2019albert} & 235M\\
            ALBERT$_{xxlarge}$ + DUMA \cite{zhu2020dual} & 292M\\
            BERT$_{large}$ \cite{devlin2019bert} & 334M\\
            XLnet$_{large}$ \cite{yang2019xlnet} & 345M\\
            RoBERTa$_{large}$ \cite{liu2019roberta} & 355M\\
            Megatron-BERT-3.9B \cite{shoeybi2019megatron} & 3.9B\\
            T5-11B \cite{raffel2020t5} & 11B\\
            T5-11B + UnifiedQA \cite{khashabi2020unifiedqa} & 11B\\
            T5-11B + UNICORN \cite{Lourie2021unicorn} & 11B\\\hline
            ALBERT$_{base}$ (rerun) & 11.2M\\
            RekNet on ALBERT$_{base}$ & 13.5M\\\hline
            ALBERT$_{xxlarge}$ (rerun) & 244M\\
            RekNet on ALBERT$_{xxlarge}$ & 277M\\\hline
		\end{tabular}
	}
\end{table}

Furthermore, we collect the training parameters of \textit{RekNet} and partial public works in Tables \ref{dream_result}-\ref{cosmos_result}, as Table \ref{parameter} shows.
In conclusion, compared with the $244M$ parameters in the baseline ALBERT$_{xxlarge}$, only approximately $13.5\%$ additional parameters are introduced into \emph{RekNet}, which is less than other comparable works \cite{zhu2020dual,zhang2020dcmn}, as well as demonstrates the conciseness and efficiency of \emph{RekNet}.

\section{Analysis}
% We adopt \textit{RekNet} on ALBERT$_{base}$ for the following analysis, and all the analyses are based on DREAM benchmark.

\subsection{Ablation Studies}
As Figure \ref{model} shows, there are two main modules (\textit{Reference Finder} and \textit{Knowledge Adapter}) in \emph{RekNet}.
To explore the necessity of each module, we remove one of them for each time and keep the hyper-parameters unchanged, obtaining the results of ablation studies on DREAM in Table \ref{ablation_result}.

\begin{table}[htbp]
\caption{\label{ablation_result} Results of ablation experiments.}
	\centering{
		\begin{tabular}{p{4cm}|l|l}
		    \hline	\bf Model & \bf Dev & \bf Test  \\
			\hline \hline
            Baseline (ALBERT$_{base}$)  & 65.74 & 65.56\\
            RekNet & 68.87 & 68.74\\
            \quad - Reference Finder & 67.72 & 67.76\\
            \quad - Knowledge Adapter & 67.65 & 67.86\\
            Relocation RekNet & 67.94 & 68.45\\\hline
		\end{tabular}
	}
\end{table}

Above results show that, removing \textit{Reference Finder} brings $1.07\%$ average performance reduction to the intact model and removing \textit{Knowledge Adapter} brings $1.05\%$ performance reduction, indicating both of proposed modules are indispensable to \emph{RekNet}.
This finding further demonstrates the rationality of the reading strategy \emph{RekNet} follows:

i) Without \textit{Reference Finder}, \textit{Knowledge Adapter} executes extensive reading blindly in full-text range, and irrelevant knowledge noise may be quoted into \emph{RekNet}, making it suffer from lacking \emph{reading} process.
However, with the well-designed knowledge structure (quadruple) as well as the numeric restriction, \emph{RekNet} can still get relatively positive improvement by quoting explicit knowledge.

ii) Without \textit{Knowledge Adapter}, \emph{RekNet} cannot introduce explicit knowledge separated from the context, leading to the lack of relevant external information and cannot reflect human \emph{comprehension} process.

To study the influence of the fusion order, we exchange the order of $RV_i$ and $KV_i$ integrates with $PV_i$, which could be considered to exchange the positions of two \textit{Integrators}.
As Table \ref{ablation_result} shows, relocation of these two modules brings negative impact to \emph{RekNet}, which may because:

i) The context in \emph{Reference Span} has a higher degree of similarity to the initial context in passage, and knowledge quadruples are extracted according to \emph{Reference Span}.
Integration of two representations of more similar contexts can makes model learn more features than combining two more different ones.

ii) \textit{Relocation RekNet} violates the order of natural human reading comprehension strategy, which may quote and over-analyze irrelevant knowledge items.

\subsection{Studies on Enriched Question Q'}
\begin{table}[htbp]
	\caption{\label{enriched_statistics} Statistics on Experimental Datasets for Q'.}
	\centering{
		\begin{tabular}{l|c|c|c}
		    \hline	\bf Dataset & \bf DREAM & \bf RACE & \bf Cosmos QA \\
			\hline \hline
			Co-occurrence Proportion (\%) & 62.8 & 55.8 & 48.1\\
			Average Length of Added Tokens & 1.91 & 2.64 & 2.44\\
			\hline
		\end{tabular}
	}
\end{table}

To show the advantages of $Q'$ compared with $Q$, we show more detailed examples in DREAM and RACE in Table \ref{more_enriched}. 
We present comprehensive statistics on the proportion of co-occurrence examples on all experimental datasets, together with the average length of added tokens, as Table \ref{enriched_statistics} shows.
Most original questions will be enriched by co-occurrence information with brief but critical words according to the limited length of added tokens, avoiding excessive redundant words, such as Dialogue 2 and Document 1 in Table \ref{more_enriched}.

\begin{table}[htbp]
\caption{\label{more_enriched} More sample dialogues and documents in DREAM and RACE for $Q'$, where \textbf{bold texts} in passage represent critical information for answer prediction while those in answer candidates represent co-occurrence information for the generation of $Q'$.}
\centering
\begin{tabular}{|p{8.2cm}|}
\hline \bf Dialogue 2\\\hline
...\\
\emph{M: We'll take the two rooms.}\\
\emph{W: Very good, sir. Would you please \textbf{register}? \textbf{Write your name and address} on this card. Thank you. Is this your luggage?}\\
\emph{M: Yes. We have four suitcases.}\\
\emph{W: All right. The bellboy will bring them up for you. You will be in rooms 403 and 405. How long do you plan to stay in Boston?}\\
...\\\hline
Q: \emph{The man has to \_\_\_.}\\
A. \emph{\textbf{register} by writing his name and address.} (correct)\\
B. \emph{\textbf{register} for his suitcases.}\\
C. \emph{\textbf{register} for the tour arrangement.}\\\hdashline
Q': \emph{Register. The man has to \_\_\_.}\\\hline
\bf Document 1\\\hline
... \emph{But \textbf{early-airport people} get ulcers, heart attacks and bite their fingernails to the bone.
\textbf{Late-airport people} almost don't realize they are flying.
A guy of that kind once said, \textbf{don't hurry}.
If you miss your flight, it's because God doesn't want you to go.} ...\\\hline
Q: \emph{We can learn from the passage that \_\_\_.}\\
A. \emph{late-\textbf{airport persons} are often nervous.}\\
B. \emph{early-\textbf{airport persons} are always at ease during the flight.}\\
C. \emph{early-\textbf{airport persons} get their baggage first after the landing.}\\
D. \emph{late-\textbf{airport persons} always take things easy.} (correct)\\\hdashline
Q': \emph{Airport person. We can learn from the passage that \_\_\_.}\\\hline
\end{tabular}
\end{table}

Furthermore, to learn the improvement brought by enriched information, we design a degradation experiment on DREAM, by degenerate $Q'$ to $Q$ in \emph{RekNet}.
Results in Table \ref{enriched_result} indicate that $Q'$ can improve the performance of MRC model and the main contribution of $Q'$ is to help model get more precise \emph{Reference Spans} for multi-choice MRC tasks.
Besides, to expand \emph{RekNet} to other MRC categories such as extractive MRC, $Q'$ can be degenerated into $Q$, with acceptable performance degradation.

\begin{table}[htbp]
\caption{\label{enriched_result} Results of question degradation experiments.}
	\centering{
		\begin{tabular}{p{5.5cm}|l|l}
		    \hline	\bf Q or Q' & \bf Dev & \bf Test  \\
			\hline \hline
			Q' to All Modules & 68.87 & 68.74 \\
			Q to All Modules & 67.60 & 68.12 \\
			Q' to Reference Finder and Q to Other Modules & 68.46 & 68.50 \\
			\hline
		\end{tabular}
	}
\end{table}

\subsection{Studies on Reference Span}
In this section, we conduct a series of experiments to explore the effect of \textit{Reference Span} on \textit{RekNet}, in terms of \textbf{quality}, \textbf{quantity} and \textbf{integration method}.

i) \textbf{Quality of \textit{Reference Span}.}
Since the framework of \textit{RekNet} employs a two-stage design (\textit{Reference Span} extraction at the first stage and knowledge quadruple injection at the second stage), one natural concern is that, whether the quality of \textit{Reference Span} has a significant impact on the ultimate performance of \textit{RekNet}, especially when the quality of \textit{Reference Span} is relatively low.
Therefore, we design three baseline models with different \textit{Reference Spans} in lower quality than the original \textit{RekNet} adopted:
\begin{itemize}
\item \textbf{TF-IDF Reference Span}: A heuristic TF-IDF method is employed to extract \textit{Reference Span}, instead of the pre-trained \textit{Reference Extractor}.
In detail, we divide the original context into clauses\footnote{To divide the context into fragments as similar as possible to the original \textit{Reference Span}, we analyze the original \textit{Reference Span} and find that a considerable proportion is a clause divided by pause punctuation such as ``,”, ``-” and so on.}, then calculate the TF-IDF similarity score of each clause and enriched question $Q'$.
The clause with the highest score is extracted as the \textit{TF-IDF Reference Span}.

\item \textbf{Attention Reference Span}: We replace the \textit{Reference Extractor} by a simple layer of attention calculation.
In detail, we employ an untrained contextualized encoder ALBERT$_{base}$ to encode each clause and enriched question $Q'$, then computes the attention score of the embedding of $Q'$ and each clause:
$$s_i=emb_{cla_{i}}^\top emb_{que}, i\in (1,...,N_{cla}),$$
where $emb_{cla_{i}}, emb_{que}$ are the embeddings of the $i$-th clause and enriched question $Q'$\footnote{Following single sentence setting in \cite{devlin2019bert}, the input sequences are ``$[CLS]\;cla_{i}$” and ``$[CLS]\;Q'$” respectively, and $emb_{cla_{i}}, emb_{que}$ are the embedding vectors of $[CLS]$ token.}, $N_{cla}$ is the number of clause, and $s_i$ is the attention score of the $i$-th clause.
Then the clause with the highest score $s_i$ is extracted as the \textit{Attention Reference Span}.

\item \textbf{Weak Reference Span}: For the low-quality \textit{Reference Span} in this baseline, we pre-train the \textit{Reference Extractor} on SQuAD 2.0 with only one training epoch to make the pre-training process inadequate\footnote{As a result, the performance of Exact Match (EM) on SQuAD 2.0 drops from 79.21 (2 training epochs, for the original \textit{Reference Extractor}) to 73.17 (1 training epoch).}, and keep other processes and settings unchanged.
\end{itemize}

ii) \textbf{Quantity of \textit{Reference Span}.}
To study whether extraction of multiple \textit{Reference Spans} can bring positive benefits to \textit{RekNet}, we design a baseline which lets \textit{Reference Extractor} extract multiple \textit{Reference Spans}, compared with one single \textit{Reference Span} in proposed \textit{RekNet}.
First, we set the probability score of the \textit{Reference Span} in the original \textit{RekNet} as $s_{max}$, which is scored by \textit{Reference Extractor}.
Then in \textit{RekNet} with multiple \textit{Reference Spans}, we modify \textit{Reference Extractor} to retain all \textit{Reference Spans} whose probability scores $s$ satisfying: $s \ge 0.7 \times s_{max}$, to build up \textit{Reference Span Set}.
Ultimately, we remove overlapping sub-spans and splice all elements in \textit{Reference Span Set} with ``.”, and replace the original \textit{Reference Span} with it.

iii) \textbf{Integration Method of \textit{Reference Span}.}
In our experiments, for each $1,000$ training steps, the baseline ALBERT costs $134/1,267$ seconds on $base$ and $xxlarge$ size respectively, while the original \emph{RekNet} costs $196/2,063$ seconds respectively, with $54.5\%$ additional training time cost on average.
Though the proportion of additional training time cost is not at a low level like additional parameters, we find that most of the additional training cost (more than $70\%$) comes from the encoding of \textit{Reference Span}.

Thus, we explore one simplified integration method of \textit{Reference Span} to simplify the integration of \textit{Passage Vector} $PV$ and \textit{Reference Vector} $RV$, and improve the overall training speed of \emph{RekNet}.
In detail, we design an additional feature encoding layer in the \textit{Encoder} in Figure \ref{model}, to encode the original context with \textit{Reference Span} as one additional indicating feature.
Since the integrated embedding $E$ encodes both the original passage and \textit{Reference Span}, we assign $E$ to both $PV$ and $RV$ in \emph{RekNet}, and remove \textit{Reference Encoder} in this baseline.

For the detailed design of above simplified integration method, the integrated embedding in above baseline is the normalized sum of the original \textit{Contextualized Embedding} and proposed \textit{Reference Embedding}.
Following the basic design in embedding layers of BERT-style models, the original \textit{Contextualized Embedding} consists of Token Embedding $E_t$, Segmentation Embedding $E_s$ and Position Embedding $E_p$.
For \textit{Reference Embedding} $E_{Ref}$, we implement another embedding layer, which has $1$ (for tokens in \textit{Reference Span}) and $0$ (for other tokens) two possible assignment values.
The proposed \textit{Reference Embedding} has the same embedding size as \textit{Contextualized Embedding}, guaranteeing all above indicating embeddings are in the same vector space.
Formulaically, the integrated embedding $E$ can be represented as:
$$E=Norm(E_t+E_s+E_p+E_{Ref}),$$
where $Norm()$ is a layer normalization function \cite{ba2016norm}.

\begin{table}[htbp]
\caption{\label{multiple_result} Results of \textit{RekNet} and baselines about \textit{Reference Span} in $base$ size on DREAM.}
	\centering{
		\begin{tabular}{l|l|l}
		    \hline	\bf Model & \bf Dev & \bf Test  \\
			\hline \hline
			Baseline (ALBERT$_{base}$) & 65.74 & 65.56\\
            RekNet & 68.87 & 68.74\\\hdashline
            \quad + TF-IDF Reference Finder & 68.04 & 68.19\\
            \quad + Attention Reference Finder & 68.39 & 68.35\\
            \quad + Weak Reference Finder & 68.63 & 68.48\\\hdashline
            \quad + Multiple Reference Finder & 68.01 & 68.28\\
            \quad + Simplified Integration & 68.45 & 68.40\\\hline
			\hline
		\end{tabular}
	}
\end{table}

Retaining all other settings the same, we evaluate \textit{RekNet} and above baselines about \textit{Reference Span} in $base$ size on DREAM, and the results are shown in Table \ref{multiple_result}, which indicate that:

i) \textbf{Quality of \textit{Reference Span}.}
Though the overall performance of \textit{RekNet} with a low-quality \textit{Reference Span} is lower than the original \textit{RekNet}, the slight degree of reduction illustrates that, \textit{RekNet} does not suffer a lot from low-quality \textit{Reference Span} like most two-stage framework models.
\textit{RekNet} possesses satisfactory robustness, especially when the extraction of \textit{Reference Span} is not completely accurate.

The robustness may benefit from the integration with Passage Vector $PV$, which can be seen as a protective measure to the low-quality \textit{Reference Span}.
Besides, knowledge quadruple injection at the second stage does not have a strict requirement on the accuracy of \textit{Reference Span}, since the \textit{Reference Span} only serves as a guild for the contribution proportion of knowledge quadruples.
One rough or even inaccurate range with less computation cost for \textit{Reference Span} can still work well for the overall performance of \textit{RekNet}.
% In addition, we find that the performance of above different low-quality \textit{Reference Spans} satisfying: 
% $$Weak Reference Finder > Attention Reference Finder > TF-IDF Reference Finder,$$
% indicating the positive influence of more parameters in \emph{Reference Finder}.

ii) \textbf{Quantity of \textit{Reference Span}.}
We find the performance of \textit{RekNet} gets a slight negative impact when it extracts multiple \textit{Reference Spans}.
The possible reasons may be, with multiple \textit{Reference Spans}, model will distract attention to more irrelevant \textit{Reference Spans}.
Furthermore, more knowledge quadruples with weak correlation (in weak-relevant \textit{Reference Spans}) will be quoted into \textit{RekNet}, leading to further negative impact.

iii) \textbf{Integration Method of \textit{Reference Span}.}
With above simplified integration method, though the overall performance of \emph{RekNet} drops slightly ($0.38\%$ on $base$ size on average), the training time of \emph{RekNet} reduces to $147/1,542$ seconds on $base$ and $xxlarge$ size for $1,000$ training steps, with $15.7\%$ additional cost on average.
Thus, we recommend \textit{RekNet} with this simplified integration method to the researchers who pursue higher training or inference speed.

\subsection{Analysis of Knowledge Quadruples}
Instead of knowledge triplets in existing studies \cite{mihaylov2018knowreader,lin2019kagnet,feng2020mhgrn}, \emph{RekNet} quotes and encodes explicit knowledge quantitatively in the form of quadruples and achieves significant improvement.
To prove the effectiveness of knowledge quadruples we proposed, we degenerate knowledge quadruples to knowledge triplets as baseline, by replacing the confidence values in knowledge quadruples with the values of $0-1$ Mask Vectors.

In detail, Mask Vector is in the vector space of $\mathbb{R}^{k\times n}$ for each $(R,A)$ binary.
We denote Mask Vector as $MV$, where $MV_i$ is the $i$-th element of $MV$.
Then we set:
$$
\label{masked_formula}
MV_i=\left\{
\begin{array}{rcl}
1 & & {c_i > \beta,}\\
0 & & {Otherwise.}
\end{array} \right.
$$
\noindent for Mask Vector, where $\beta$ is a non-negative threshold.

\begin{table}[htbp]
\caption{\label{degeneration_result} Result of knowledge degeneration experiments.}
	\centering{
		\begin{tabular}{p{4cm}|l|l}
		    \hline	\bf Model & \bf Dev & \bf Test  \\
			\hline \hline
            Baseline (ALBERT$_{base}$)  & 65.74 & 65.56\\
            RekNet & 68.87 & 68.74\\
            Masked RekNet ($\beta = 0$) & 68.28 & 68.05\\\hline
		\end{tabular}
	}
\end{table}

We first set $\beta = 0$ to degenerate knowledge quadruples into triplets, which treats all knowledge items equally. 
Result in Table \ref{degeneration_result} indicates that, Mask Vector performs worse than the intact \textit{RekNet} with confidence values, because there exists a huge amount of untrustworthy knowledge items in knowledge graph (ConceptNet), such as \emph{(abdomen, relatedto, thorax, 0.102)}.
It is unreasonable to give them equal treatment to trustworthy ones.

\begin{figure}[htbp]
\centering
\includegraphics[scale=0.6]{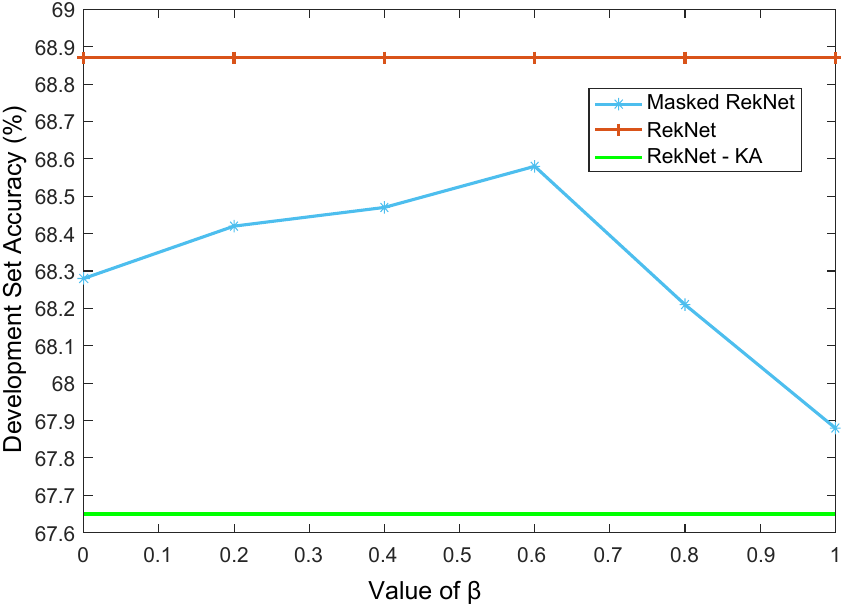}
\caption{The accuracy of \emph{Masked RekNet} with different thresholds ($\beta$) on development set. KA: Knowledge Adapter.}
\label{mask_result}
\end{figure}

Then we implements masked experiments over different $\beta$ on the development set of DREAM, and the results are shown in Figure \ref{mask_result}.
The figure shows that, among different thresholds, \emph{Masked RekNet} performs better than \emph{RekNet} without \textit{Knowledge Adapter} (no knowledge), but worse than the intact \emph{RekNet} (with knowledge quadruples).

Furthermore, the reasons for the trend of model performance with $\beta$ may be that, with lower $\beta$, model cannot filter out untrustworthy knowledge items effectively, while with higher $\beta$ model may filter out some credible and important knowledge.
In summary, the results of above masked experiments prove the effectiveness of confidence values.

\begin{figure}[htbp]
\centering
\includegraphics[scale=0.9]{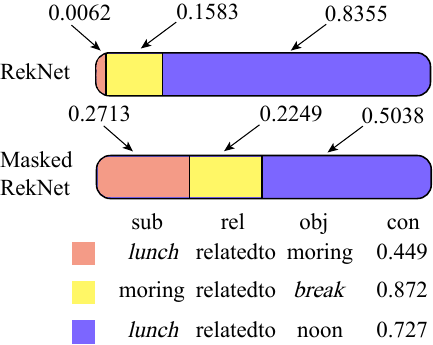}
\caption{The visual contribution proportion of each knowledge item for the dialogue in Figure \ref{example}. In this example, red, yellow and blue knowledge items are untrustworthy, irrelevant and golden ones, respectively.}
\label{visiable}
\end{figure}

To show the effectiveness of knowledge filtering method of \emph{RekNet} (using knowledge quadruples) in a visual way, we extract the final contribution proportion of each knowledge item for the example in Figure \ref{example}, and display the visualized proportions in Figure \ref{visiable}, together with the proportions in \emph{Masked RekNet} (using mainstream knowledge triplets).

As the figure depicts, by adopting knowledge quadruples, \emph{RekNet} gives the lowest proportion to \emph{(lunch, relatedto, morning, 0.449)} because it is untrustworthy, and gives \emph{(morning, relatedto, break, 0.872)} a relatively low proportion in spite of its high confidence value, because it has less contextual relevance to the \emph{Reference Span}.
Compared with the intact \emph{RekNet}, \emph{Masked RekNet} adopting knowledge triplets may unnecessarily pay attention to untrustworthy knowledge items and less attention to precise ones, leading to wrong predictions.

\subsection{Statistics about Retrieved Knowledge}
In order to provide researchers with more useful information about the utilization of external knowledge in multi-choice MRC tasks, we make a brief statistics about the retrieved knowledge by \emph{RekNet} to explore:

i) the overall demand of questions in multi-choice MRC tasks;

ii) which type of knowledge contributes more to multi-choice MRC tasks, or tends to attract more attention from proposed model.

As a result, we find $77.82\%$, $72.55\%$ and $84.73\%$ questions in DREAM, RACE and CosmosQA are covered by knowledge respectively, and each \textbf{covered} $(Q,A_i)$ pair has $2.42, 3.90, 2.69$ knowledge items on average ($3.08$ quadruples in total).
And for all questions (including uncovered questions), each question retrieves $9.84$ knowledge quadruples on average ($2.49$ quadruples for each $(Q,A_i)$ pair).

Above statistics proves the rationality of our setting of $k=5$ (the number of knowledge quadruples for each candidate answer) in the setup of experiments.
Due to each $(Q,A_i)$ pair covered by knowledge retrieves approximately $3$ pieces of knowledge items on average (nearly $4$ pieces for RACE), smaller \emph{k} may lead to the loss of potentially important knowledge, while larger \emph{k} might bring knowledge noise and unnecessary computational consumption.
When $k = 4$ and $6$, the average performance on ALBERT$_{base}$ drops by $0.24\%$ and $0.19\%$ on RACE compared with $k=5$.

As for the relations in knowledge quadruples, we have $20$ types of different relations in total, and top $3$ relationships in the original ConceptNet are: ``relatedto” ($67.65\%$), ``isa” ($17.48\%$) and ``situation” ($7.12\%$), but the top $3$ relationships retrieved by \emph{RekNet} are ``relatedto” ($72.13\%$), ``antonym” ($7.56\%$) and ``locate” ($5.84\%$). 
This finding indicates that:

i) \emph{RekNet} tends to retrieve ``relatedto” relationship due to this type of knowledge items accounts for more than two thirds knowledge items in the original ConceptNet; 

ii) Knowledge item which indicates negative meaning (antonym: $0.72\% \rightarrow 7.56\%$) or is hard to learn by contextualized encoder (locate: $0.90\% \rightarrow 5.84\%$) is helpful for \emph{RekNet}, instead of similar entities which can be easily learned by models (isa: $17.48\% \rightarrow 5.68\%$).

\subsection{Expand on Extractive MRC}
In addition to the multi-choice MRC tasks focused on by this work, we further expand \textit{RekNet} to other MRC categories such as extractive MRC tasks, to verify the versatility of the proposed methods in \textit{RekNet}.

Following \cite{devlin2019bert}, the input sequence for extractive MRC can be represented as $[CLS]\;Q\;[SEP]\;P\;[SEP]$.
We define the representation on the final hidden layer for the $i$-th input token as $T_i \in \mathbb{R}^H$, then we score the answer span starts from the $i$-th input token and ends at the $j$-th input token as $S \cdot T_i+E \cdot T_j$, where $S/E\in \mathbb{R}^H$ is the introduced start/end vector. 
Ultimately, span with the largest score of the former formula is chosen as the predicted answer span.
Denoting $SP_i=\frac{e^{S \cdot T_i}}{\sum_j e^{S \cdot T_j}}$ as the probability of the $i$-th token being the start of the answer span and $EP_i=\frac{e^{E \cdot T_i}}{\sum_j e^{E \cdot T_j}}$ as the probability of being the end of the answer span, our training object is the sum of Cross Entropy Losses for the start and end token probabilities.

And for the implementation of \textit{RekNet} on extractive MRC, we design the following adjustments to bridge the gap from multi-choice MRC:

i) As we mention in Section VI.B, we degenerate $Q'$ into $Q$ for the input sequences of \textit{RekNet}, due to no candidate answer being given.

ii) For the representation alignment of each token in extractive MRC task, we remove \textit{Reference Encoder} module and adopt the simplified integration method (indicator embedding layer for \emph{Reference Span}, mentioned in Section VI.C) to integrate the given passage and extracted \emph{Reference Span}.

iii) \emph{Knowledge Finder} searches question-relevant knowledge items based on $(R,Q)$ binary instead of $(R,A)$ binary in multi-choice MRC, and we set the maximum number $m$ of knowledge quadruples as $m=k=5$.

iv) We enhance the embedding vector of each input token by the integration of \textit{Knowledge Vector}.
Similar to the \textit{Integrator} in Section IV.C, we execute concatenation and dimensionality reduction operations for them.

We evaluate above adjusted \textit{RekNet} on ALBERT$_{base}$ on the development set of SQuAD 1.1 \cite{rajpurkar2016squad}, and provide an example to show how \textit{RekNet} works on extractive MRC tasks in Table \ref{squad}.
Corresponding ablation experiment results are shown in Table \ref{extractive_result}.

\begin{table}[htbp]
\caption{\label{squad} Sample document in SQuAD 1.1, where \textbf{bold texts} represent \emph{Reference Span} and knowledge quadruples are in the form of ``(subject, relation, object, confidence value) $\rightarrow$ contribution proportion”.}
\centering
\begin{tabular}{|p{8.2cm}|}
\hline \bf Document 2 \\ \hline
... \emph{Fox paid for \textbf{Deadpool, X-Men: Apocalypse, Independence Day: Resurgence and Eddie the Eagle}, Lionsgate paid for Gods of Egypt, Paramount paid for Teenage Mutant Ninja Turtles: Out of the Shadows and 10 Cloverfield Lane,} ...\\\hline
Q: \emph{What famous July Fourth holiday movie did Fox pay to advertise a sequel of during the Super Bowl?}\\\hdashline
Knowledge Quadruples:\\
\emph{(independence, relatedto, july\_fourth, 0.591) $\rightarrow$ 0.5697}\\
\emph{(independence, relatedto, july, 1.297) $\rightarrow$ 0.2688}\\
\emph{(independence\_day, isa, holiday, 1.000) $\rightarrow$ 0.1501}\\
\emph{(holiday, isa, day, 2.000) $\rightarrow$ 0.0103}\qquad\emph{(x, isa, movie, 1.000) $\rightarrow$ 0.0011}\\\hdashline
A: \emph{Independence Day: Resurgence} (correct answer)\\\hline
\end{tabular}
\end{table}

\begin{table}[htbp]
\caption{\label{extractive_result} Ablation experiment results of \textit{RekNet} for extractive MRC.}
	\centering{
		\begin{tabular}{p{4cm}|c|c}
		    \hline	\bf Model & \bf EM & \bf F1  \\
			\hline \hline
            Baseline (ALBERT$_{base}$) & 82.66 & 89.91\\
            Adjusted RekNet & 84.03 & 90.98\\
            \quad - Reference Finder & 82.97 & 90.42\\
            \quad - Knowledge Adapter & 83.58 & 90.69\\\hline
		\end{tabular}
	}
\end{table}

As Document 2 in Table \ref{squad} shows, in the adjusted \textit{RekNet}, \textit{Reference Finder} extracts one less precise span as \emph{Reference Span} (e.g., the bold texts in Document 2).
In the above example, all movies paid by Fox have been extracted coarsely due to the lack of commonsense about \textit{``July Fourth holiday”} in the question.
Then \textit{Knowledge Adapter} quotes related knowledge quadruples and scores them with final contribution proportions, highlighting critical information \textit{``Independence Day”} in the \emph{Reference Span}.
With the integration of the above information, our adjusted \textit{RekNet} ultimately enhances the embedding vector of each token and predicts the precise answer span.

Furthermore, as Table \ref{extractive_result} shows, though some benefit components or methods are discarded for extractive MRC (such as enriched question $Q'$ and \textit{Reference Encoder}), the improvements in above ablation results prove that, \textit{RekNet} possesses satisfactory versatility on diverse MRC categories such as extractive MRC.
Besides, the limited contribution of \textit{Knowledge Adapter} on extractive MRC indicates that, in most cases extractive MRC may benefit less from external knowledge injection than multi-choice MRC, which may be one main reason that most existing researches about knowledge enhancement \cite{shwartz2020unsupervised,xia2019auxiliary,lin2021differentiable,lin2019kagnet,feng2020mhgrn,lv2020knowledge} as well as this work focus on multi-choice MRC tasks, instead of other MRC categories.

\subsection{Error Case Analysis}
\begin{table}[htbp]
\caption{\label{dialogue1} Sample dialogue in DREAM requiring numeral calculation, where \textbf{bold texts} represent critical information which needs numerical calculation for answer prediction.}
\centering
\begin{tabular}{|p{8.2cm}|}
\hline \bf Dialogue 3 \\ \hline
...\\
\emph{W: The price for one person for a ten-day tour is only \textbf{\$1,088}, which includes round-trip airfare.}\\
\emph{M: That sounds reasonable. By the way, do you have a discount for two?}\\
\emph{W: Yes, you can have \textbf{a 10\% discount}.}\\\hline
Q: \emph{If \textbf{the man and his wife} go on the recommended package tour, how much should they pay?}\\
A. \emph{\$1,088.}\\
B. \emph{\$1,958.} (correct)\\
C. \emph{\$2,176.}\\\hline
\end{tabular}
\end{table}

We extracte $50$ error cases of \emph{RekNet} based on ALBERT$_{base}$ on DREAM randomly, finding $36\%$ of them are related to logical calculation (especially numerical calculation), as Table \ref{dialogue1} shows.
On the contrary, the original DREAM, RACE and Cosmos QA only have $14\%$, $8\%$ and $4\%$ examples requiring logical calculation respectively.
Though \emph{Reference Span} in \emph{RekNet} can reveal computing words like \emph{discount}, \emph{half} from passages and questions, it fails to calculate the correct result due to the lack of human numeral logical operation, which calls for more in-depth researches in MRC field.

\section{Conclusion}
To alleviate the challenge of knowledge role missing in multi-choice MRC, this work makes the first attempt to integrate \emph{explicit knowledge} based on \emph{reference span extraction} into MRC modeling, presenting \emph{\textbf{Re}ference \textbf{K}nowledgeable \textbf{Net}work (RekNet)}, which can quote relevant and credible explicit knowledge for multi-choice MRC tasks.
We evaluate the proposed \emph{RekNet} on three typical multi-choice MRC benchmarks: RACE, DREAM and Cosmos QA, and obtain consistent and significant performance improvements which pass the significance tests.
In the future, we will expand other common reading strategies to \emph{RekNet}.

\ifCLASSOPTIONcaptionsoff
  \newpage
\fi

\bibliographystyle{IEEEtran}
\bibliography{reknet}

\end{document}